# Exploring Semi-supervised Variational Autoencoders for Biomedical Relation Extraction

Yijia Zhang[a,b] and Zhiyong Lu[a*]

[a] National Center for Biotechnology Information (NCBI), National Library of Medicine (NLM), National Institutes of Health (NIH), Bethesda, Maryland 20894, USA

[b] School of Computer Science and Technology, Dalian University of Technology, Dalian, Liaoning 116023, China

**Corresponding author: Zhiyong Lu (zhiyong.lu@nih.gov)**

**Abstract**

The biomedical literature provides a rich source of knowledge such as protein-protein interactions (PPIs), drug-drug interactions (DDIs) and chemical-protein interactions (CPIs). Biomedical relation extraction aims to automatically extract biomedical relations from biomedical text for various biomedical research. State-of-the-art methods for biomedical relation extraction are primarily based on supervised machine learning and therefore depend on (sufficient) labeled data. However, creating large sets of training data is prohibitively expensive and labor-intensive, especially so in biomedicine as domain knowledge is required. In contrast, there is a large amount of unlabeled biomedical text available in PubMed. Hence, computational methods capable of employing unlabeled data to reduce the burden of manual annotation are of particular interest in biomedical relation extraction. We present a novel semi-supervised approach based on variational autoencoder (VAE) for biomedical relation extraction. Our model consists of the following three parts, a classifier, an encoder and a decoder. The classifier is implemented using multi-layer convolutional neural networks (CNNs), and the encoder and decoder are implemented using both bidirectional long short-term memory networks (Bi-LSTMs) and CNNs, respectively. The semi-supervised mechanism allows our model to learn features from both the labeled and unlabeled data. We evaluate our method on multiple public PPI, DDI and CPI corpora. Experimental results show that our method effectively exploits the unlabeled data to improve the performance and reduce the dependence on labeled data. To our best knowledge, this is the first semi-supervised VAE-based method for (biomedical) relation extraction. Our results suggest that exploiting such unlabeled data can be greatly beneficial to improved performance in various biomedical relation extraction, especially when only limited labeled data (e.g. 2000 samples or less) is available in such tasks.

# 1. Introduction

Currently there are over 28 million articles in PubMed, and each year the biomedical literature grows by more than one million articles [1, 2]. As a result, a vast amount of valuable knowledge about proteins, drugs, diseases and chemicals, critical for various biomedical research studies, is "locked" in the unstructured free text [3, 4]. With the rapid growth, it is increasingly challenging to manually curate information from biomedical literature, such as protein-protein interactions (PPIs), drug-drug interactions (DDIs) and chemical-protein interactions (CPIs). The goal of information extraction in biomedicine is to automatically extract biomedical relations through advanced natural language processing (NLP) and machine learning techniques.

Over the past decade, a number of hand-annotated datasets have been created for biomedical relation extraction such as the various PPI corpora [5] and DDI 2013 corpus [6]. Based on these public corpora, a number of methods [7-10] have been attempted. For example, Airola, et al. (2008) proposed an all path kernel approach to extract PPIs based on the lexical and syntactic features from the dependency syntactic graph. Zhang, et al. (2012) proposed a hash subgraph pairwise kernel method to extract DDIs, which efficiently generates hash features from a dependency syntactic graph based on the hash operation. More recently, models based on deep neural networks such as convolutional neural networks (CNNs) and recurrent neural networks (RNNs), have shown promising results in various tasks [11-13].

Most of the aforementioned high-performing systems in biomedical relation extraction to date are based on supervised machine-learning approaches, which are known to be dependent on manually labeled data. Although data exist for few relation types such as PPIs and DDIs, there is a lack of large-scale training data for many other critical relationships in the biomedical domain, including gene-disease, drug-disease, and drug-mutation, due to the prohibitive expense of manual annotation. In response, alternative methods such as distant supervision have been proposed [14, 15], which automatically create training data based on existing curated biological databases. However, since the databases are generally incomplete and our language is highly rich and diverse, such automatically created labeled data are always noisy [16].

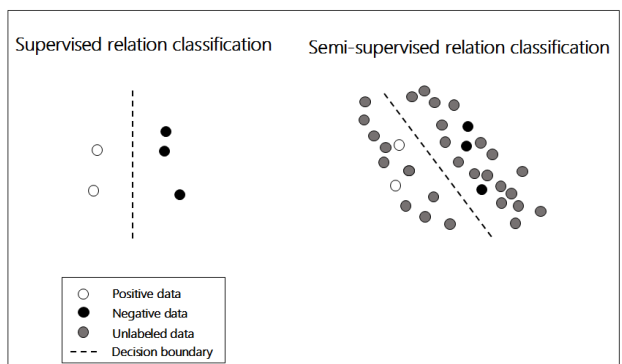

**Fig. 1. An illustrative example of a semi-supervised learning method.** The white, black and gray circles represent positive, negative and unlabeled relation instances, respectively. The dashed line represents the decision boundary decided by the classifier.

In this paper, we introduce a novel semi-supervised variational autoencoder (VAE)-based method for biomedical relation extraction. In contrast to supervised learning, the semi-supervised method learns discriminative features from both labeled and unlabeled data. Fig. 1 represents an illustrative example of the difference between supervised and semi-supervised learning methods. It is difficult to predict the decision boundary accurately based on a small number of labeled instances. However, if we integrate the unlabeled instances – generally distributed according to a mixture of individual-class distributions – with the labeled ones, we may considerably improve the learning accuracy. Given the great abundance of freely available texts in PubMed, it is particularly valuable to explore semi-supervised-based methods for improving biomedical relation extraction.

Motivated by the recent success of semi-supervised VAE methods in image classification [17, 18], text modeling [19] and text classification tasks [20], in this work we investigate its feasibility for relation extraction, which differs significantly from the other tasks. To the best of our knowledge, this is the first VAE-based method for (biomedical) relation extraction. To demonstrate its robustness, we validate our method on multiple different biomedical relation types: PPIs, DDIs and CPI extraction.

## 2. Materials and methods

### 2.1 Biomedical Relation Extraction

Biomedical relation extraction is generally approached as the task of classifying whether a specified semantic relation holds between two biomedical entities within a sentence or document. According to the number of semantic relation classes, biomedical relation extraction can be further categorized into binary vs. multi-class relation extraction.

In this paper, we focus on PPI, DDI and CPI extraction. PPI is a binary relation extraction task, whereas DDI and CPI are multi-class relation extraction task. We show some examples as follows.

- **PPI extraction example:** *These results suggest that profilin may be involved in the pathogenesis of glomerulonephritis by reorganizing the actin cytoskeleton.*

- **DDI extraction example:** *The concomitant administration of gemfibrozil with Targretin capsules is not recommended.*

- **CPI extraction example**: *Compound C diminished AMPK phosphorylation and enzymatic activity, resulting in the reduced phosphorylation of its target acetyl CoA carboxylase.*

In the case of PPIs, a system only needs to identify whether the candidate entity pair has a semantic relation or not. For DDI and CPI, a system requires not only the detection of the semantic relation between two candidate entities but also the classification of the specific semantic relation into the correct type. For example, the DDI extraction task requires a system to distinguish five different DDI types, including *Advice*, *Effect*, *Mechanism*, *Int* and *Negative*. Similarly, the CPI extraction task includes six specific types: *Activator*, *Inhibitor*, *Agonist*, *Antagonist*, *Substrate* and *Negative.*

### 2.2 Datasets

For PPI, we used the BioInfer dataset [21]. For DDI, the DDI extraction 2013 corpus [6, 22] was used. For CPI, we use the recent ChemProt corpus [23], which was served as the benchmarking data in the BioCreative VI challenge task. The detailed statistics of the datasets are listed in Tables 1, 2 and 3, respectively.

**Table 1.** The statistics of the PPI corpus

| Dataset | Sentences | Positive | Negative | Total |
|---|---|---|---|---|
| BioInfer | 1100 | 2,534 | 7,132 | 9,666 |

**Table 2.** The statistics of the DDI corpus

| Dataset | Advice | Effect | Mechanism | Int | Negative | Total |
|---|---|---|---|---|---|---|
| Training set | 826 | 1,687 | 1,319 | 188 | 23,772 | 27,792 |
| Test set | 221 | 360 | 302 | 96 | 4,737 | 5,716 |

**Table 3.** The statistics of the ChemProt corpus

| Relations | Training set | Development set | Test set |
|---|---|---|---|
| Active | 768 | 550 | 664 |
| Inhibitor | 2251 | 1092 | 1661 |
| Agonist | 170 | 116 | 194 |
| Antagonist | 234 | 197 | 281 |
| Substrate | 705 | 457 | 643 |
| Negative | 12461 | 8070 | 11013 |
| Total | 16589 | 10482 | 14456 |

## 2.3 Our Semi-supervised VAE-based Approach

### 2.3.1 Overview of our approach

A schematic overview of our method is shown in Fig. 2. Our approach includes three parts: a classifier, an encoder, and a decoder. And in a nutshell, our semi-supervised VAE model works by calibrating the classifier through the use of the encoder and decoder. First, the encoder network encodes all the labeled and unlabeled data into the latent space. Next, the semi-supervised VAE samples the latent variable $z$. Then, the decoder network strives to reconstruct the input based on the latent variable $z$ and the discrete label $y$. The latent variable distribution is usually parametrized by a Gaussian distribution, with its mean and variance generated by the encoder.

In our work, the classifier takes as input a sentence with an entity pair, where each word is represented by a word vector embedding and position embedding. The input of the decoder is a vector combined by the latent variable $z$ generated by the encoder (in blue) and the output of our classifier: the known label $y$ or the predicted label $y'$ depending on whether the input sentences are the labeled data or not (in orange). In this way, the predicted labels $y'$ of the classifier directly affect the output of the decoder. Intuitively, an accurate predicted label $y'$ will be much more informative and helpful for the decoder to reconstruct the input of the encoder than a random $y'$. Therefore, when the input sentences are the unlabeled data, the optimization of the reconstruction loss will benefit the classifier.

Based on the proposed model, the unlabeled data are incorporated with the labeled data to help train the parameters of the classifier, which means that the semi-supervised VAE model is learning useful features from both labeled and unlabeled data.

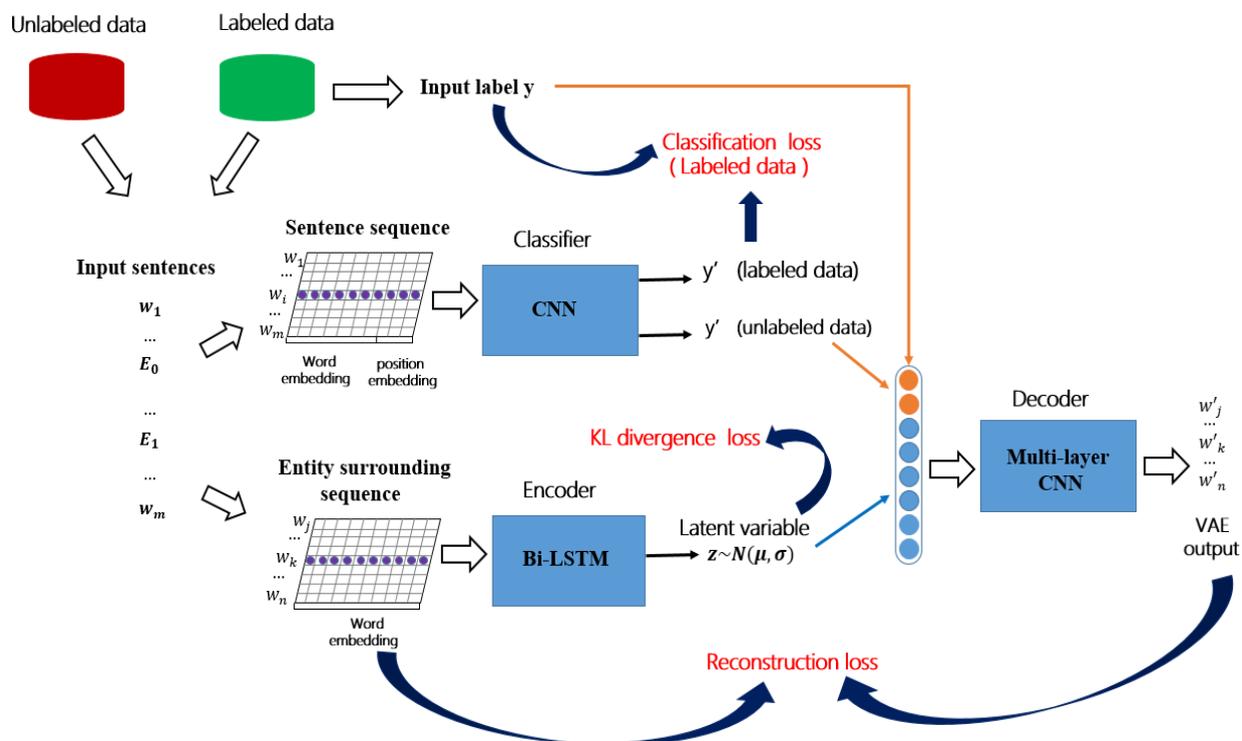

**Fig. 2.** The overview of the semi-supervised VAE-based model for relation extraction.

### 2.3.2 The semi-supervised VAE model

We briefly introduce the semi-supervised variational inference here [17]. Let $(x,y) \sim D_l$ and $x \sim D_u$ denote the labeled data and unlabeled data, respectively. The semi-supervised generative model [17] was proposed to incorporate the continuous latent variable $z$ and the discrete label $y$.

$$p(x,y,z) = p(y)p(z)P(x|y,z) \qquad (1)$$

The semi-supervised variational model consists of three parts, including a discriminative network $q(y|x)$, an inference network $q(z|x,y)$ and a generative network $p(x|y,z)$. For the labeled data $(x,y) \sim D_l$, a data point and its label are given and the variational bound is as follows:

$$\log p(x,y) \geq \mathrm{E}_{q(z|x,y)}[\log p(x|y,z) + \log p(y) + \log p(z) - \log q(z|x,y)]$$
$$= \mathrm{E}_{q(z|x,y)}[\log p(x|y,z)] - KL(q(z|x,y)||p(z)) + \log p(y)$$
$$= -\mathcal{L}(x,y) \qquad (2)$$

where $p(z)$ is the prior distribution, $q(z|x,y)$ is the learned latent posterior, the term $\mathrm{E}_{q(z|x,y)}[\log p(x|y,z)]$ is the expectation of the conditional log-likelihood on latent variable $z$, and the term $KL(q(z|x,y)||p(z))$ is the Kullabck-Leibler divergence of $p(z)$ and $q(z|x,y)$.

For the unlabeled data $x \sim D_u$, the label $y'$ is predicted by the discriminative network $q(y|x)$. Thus, the variational bound of the unlabeled data is as follows:

$$\mathrm{Log}\, p(x) \geq \mathrm{E}_{q(y,z|x)}[\log p(x|y,z) + \log p(y) + \log p(z) - \log q(y,z|x)]$$
$$= \sum_y q(y|x)(-\mathcal{L}(x,y)) + \mathcal{H}(q(y|x))$$
$$= -U(x) \qquad (3)$$

In addition, the classification loss of the labeled data is $\mathrm{E}_{(x,y) \sim D_l}[\log(q(y|x))]$. Incorporating the labeled and unlabeled terms, the objective function of the entire dataset is as follows:

$$J = \mathrm{E}_{(x,y) \sim D_l}[\mathcal{L}(x,y)] + \mathrm{E}_{(x) \sim D_u}[U(x)] + \alpha \mathrm{E}_{(x,y) \sim D_l}[\log(q(y|x))] \qquad (4)$$

where the hyperparameter $\alpha$ balances the relative weight between generative and discriminative learning.

Typically, the discriminative network $q(y|x)$, the inference network $q(z|x,y)$ and the generative network $p(x|y,z)$ are implemented by a classifier $f_{cla}(\cdot)$, an encoder network $f_{enc}(\cdot)$ and a decoder network $f_{dec}(\cdot)$, respectively. First, the encoder network encodes all the labeled and unlabeled data into the latent space. Next, the semi-supervised VAE samples the latent variable $z$. Then, the decoder network reconstructs the input based on the latent variable $z$ and the discrete label $y$. The latent variable distribution is usually parametrized by a Gaussian distribution, with its mean and variance generated by the $f_{enc}(\cdot)$.

### 2.3.3 Embedding input representation

The input of the classifier and encoder are sentence sequences and the entity surrounding sequences. Given a sentence $S$, we replace the two candidate entities with "E0" and "E1". The classifier input is the sentence sequence $\{w_1, w_2, \ldots, w_m\}$. Word embedding [24] maps words to low-dimensional vector space and preserves the syntactic and semantic information underlying the words. Position embedding is shown to be also important for relation extraction [25], which captures the distance feature between each word to the candidate entities. In our experiments, we trained word embedding using the fastText model [26] on the entire PubMed Central (PMC) Open Access subset and the

Medical Subject Headings (MeSH). The position embedding is randomly initialized following a standard normal distribution.

Let $W_{word}$ and $W_{dis}$ denote the word embedding and position embedding, respectively. For each word $w_i$ of the sentence sequence $\{w_1, w_2, \ldots, w_m\}$, the word embedding vector $w_i^{word}$ and two position vectors $w_i^{dis0}$ and $w_i^{dis1}$ can be obtained, respectively, based on $W_{word}$ and $W_{dis}$. The final embedding representation of $w_i$ is $[(w_i^{word})^T, (w_i^{dis0})^T, (w_i^{dis1})^T]$. In addition, unlike the sentence classification or text modeling task, the words around the two candidate entities are generally more valuable than other words in the sentence for the relation extraction. In particular, there are many long and complicated sentences in the biomedical literature, where only the words surrounding the targeted entities are useful to distinguish and classify the candidate relation. Therefore, we only use the word sequence surrounding the two candidate entities as the encoder input. More specifically, we empirically choose the 10 preceding words and the 5 succeeding words of the first target entity and the first 5 preceding words and the 10 succeeding words of the second entity. The selected word sequences are concatenated to the entity surrounding sequence $\{w_j, w_k, \ldots, w_n\}$, whose size is 30. If the surrounding words of the target entities are not available in the sentence, zeros are padded in the entity surrounding sequence. We only use word embeddings to represent the entity surrounding sequence. For each word $w_j$ of the entity surrounding sequence, the embedding representation is $[(w_j^{word})^T]$.

### 2.3.4 The model architecture

The semi-supervised VAE are implemented using deep neural networks. Bidirectional LSTMs (Bi-LSTMs) and CNNs are two major neural networks for the semi-supervised VAE. Some recent studies [19, 20] have suggested the (Bi-)LSTM is more suitable to be used as encoder than CNNs. In our work, we experimented with (Bi-)LSTM and CNNs on decoder and classifier, respectively. We found the CNNs was more effective and efficient on decoder and classifier than (Bi-)LSTMs. Hence, in our semi-supervised VAE architecture the encoder and decoder are implemented by bidirectional LSTMs (Bi-LSTMs) and multilayer CNNs, respectively. For the classifier, we implemented with the CNNs architecture. Next, we provide a brief introduction of the CNNs and (Bi-)LSTM architectures.

**CNN model**: The CNN model learns the features from the text by applying a convolution operation in the sentences. Given a sentence $\{w_1, w_2, \ldots, w_m\}$, the embedding input representation is represented as $\{z_1, z_2, \ldots, z_m\}$. CNNs model [27] generally contains a set of convolution filters. Each convolution filter applies convolution operations to $n$ continuous words to generate a new feature based on n-grams, which is parameterized by the weight matrix $W$. For a given window of words $z_{i:i+n-1}$, a feature $c_i$ is generated by a convolution filter as follows:

$$c_i = f(W \cdot z_{i:i+n-1} + b) \qquad (6)$$

where $f$ is a nonlinear function, and b is a bias parameter. Thus, the convolution filter is applied to the whole sentence step by step $\{z_{1:n}, z_{2:n+1}, \ldots, z_{m-n+1:m}\}$ to produce a feature map as follows:

$$\mathbf{c} = [c_1, c_2, \ldots, c_{m-n+1}] \qquad (7)$$

To choose the most salient feature, we employ the max pooling operation [28] $\hat{c} = \max\{\mathbf{c}\}$ over the feature map $\mathbf{c}$.

**LSTM model**: The LSTM model exploits the gate mechanism to learn the long-term dependency feature from the sentences. In LSTM models [29], the current hidden state $h_j$ and memory cell $c_j$, at the time step $j$, are generated based on the current input word $x_j$, the previous hidden state $h_{j-1}$ and the previous memory cell $c_{j-1}$. Specifically, the input gate $i_j$, the forget gate $f_j$, the output gate $o_j$ and the extracted feature vector $g_j$ are defined as follows.

$$i_j = sigmoid(W_i x_j + U_i h_{j-1} + b_j) \qquad (8)$$

$$f_j = sigmoid(W_f x_j + U_f h_{j-1} + b_f) \qquad (9)$$

$$o_j = sigmoid(W_o x_j + U_o h_{j-1} + b_o) \qquad (10)$$

$$g_j = tanh(W_g x_j + U_g h_{j-1} + b_g) \qquad (11)$$

where $W_*$, $U_*$ and $b_*$ are weight matrices and bias vectors. Based on Eqs. (8)-(11), the memory cell $c_j$ and hidden state $h_j$ are calculated as follows.

$$c_j = f_j \odot c_{j-1} + i_j \odot g_j \qquad (12)$$

$$h_i = o_j \odot tanh(c_{j-1}) \qquad (13)$$

where $\odot$ denotes element wise multiplication.

The Bi-LSTMs capture contextual features from the input sentence both forward and backward. Let $h_k^f$ and $h_k^b$ denote the output of the forward LSTMs and backward LSTMs. The output of the Bi-LSTMs is the concatenation $h_k = h_k^f \parallel h_k^b$.

In our model, we use Bi-LSTM to implement the encoder. As mentioned, the encoder input is the embedding representation of the entity surrounding sequences. The encoder output is the latent variables. For the classifier, we choose CNNs architecture. The classifier input is the embedding representation of the sentence sequences. The classifier output is the one-hot label vector. We employ the *Softmax* function to implement the classification based on the feature representation generated by CNNs. The decoder is implemented by three-layer CNNs. The latent variable and the one-hot label vector are combined as the decoder input. For the labeled data, the labels are already given, so we map the labels into the one-hot label vector. For the unlabeled data, the one-hot label vector is predicted by the classifier. In our experiments, we apply a full connected layer to connect the decoder input and the three-layer CNNs. Similar to machine translation, the output of the three-layer CNNs is fed into a full connected layer to implement the reconstruction of the entity surrounding sequences.

Our semi-supervised model learns the features from both the labeled and unlabeled data. According to the Eq. 4, the loss functions of the labeled and unlabeled data are different. Our semi-supervised VAE model contains three loss terms including classification loss, reconstruction loss and KL divergence loss. The classification loss is only calculated on the labeled data. The reconstruction loss is used to reconstruct the encoder input and decoder output. The KL divergence loss is the major character of a VAE model, and is used to adjust the latent variable distribution to the Gaussian distribution. For the labeled data, we compute all three loss terms: classification loss, reconstruction loss and KL divergence loss. For the unlabeled data, we only calculate the reconstruction loss and KL divergence loss. The classification loss and reconstruction loss are calculated by cross-entropy and sparse cross-entropy. The KL divergence loss is calculated in the closed form [17]. Therefore, the objective loss for the labeled and unlabeled data are $Loss_{reconstruct} + Loss_{KL} + Loss_{classify}$ and $Loss_{reconstruct} + Loss_{KL}$, respectively.

**Table 4.** The hyperparameters used in the experiments

| Hyperparameter name | Value |
|---|---|
| Word embedding dimensionality | 200 |
| Position embedding dimensionality | 20 |
| Mini batch size | 64 |
| Latent variable size | 32 |
| Encoder Bi-LSTM hidden units | 600 |
| 1st Decoder CNNs hidden units | 300 |
| 2nd Decoder CNNs hidden units | 600 |
| 3rd Decoder CNNs hidden units | 1000 |
| Classifier CNNs hidden units | 300 |
| Encoder dropout rate after embedding | 0.5 |
| Decoder dropout rate before output | 0.5 |
| Classifier dropout rate after embedding | 0.5 |
| Classifier dropout rate before output | 0.5 |

In our experiment, the proposed model is implemented by the Keras library. In each training batch, we first sample the training instances from the labeled and unlabeled data, respectively. Then, our model is trained by labeled instances and unlabeled instances, separately. Hence, the parameters of the classifier are trained by both labeled and unlabeled data. We apply the resilient mean square propagation (RMSProp) [30] to optimize the parameters of the model. We manually tuned the hyperparameters based on the system results on the validation set. The final hyperparameters used in our experiments are listed in Table 4. A dropout mechanism was applied in the encoder, decoder and classifier, to alleviate the overfitting of the neural networks model [31].

## 3. Results

### 3.1 Evaluation metrics and setups

The standard *F-score*, *precision* and *recall* are used as the evaluation metrics in our experiments. The $F_1$-score is the harmonic mean of both *precision* and *recall*, which is defined as $(2 \cdot precision \cdot recall)/(precision + recall)$. For the DDI and CPI extraction task, we compute the micro average to assess the overall performances[12, 13]

Generally, the semi-supervised experiments need labeled, unlabeled, validation and test set. For the BioInfer corpus, we randomly chose 1000 samples as the test set and another 1000 samples as the validation set (approximately

10% of total PPI data for each set). For the DDI corpus, we randomly chose 2000 training samples as the validation set. For the ChemProt corpus, we first combined the training and development sets (as shown in Table 4) and similarly chose 2000 training samples (again, approximately 10%) as the validation set. The validation set was used for optimizing the hyperparameters and the number of epochs for our model training. The held-out test set was used for evaluating the performance of our model.

### 3.2 Experimental Results

Firstly, we used the baseline CNNs model with different number of labeled samples to evaluate the supervised classifier performance on the three corpora. After splitting the validation and testing sets, the number of labeled data for BioInfer, DDI and ChemProt are 7,666, 25,792 and 25,071, respectively. We experimented with the different labeled training rate which ranges from 1% to 100% on three corpora. For BioInfer, the labeled training rate ranges from 1% (77) to 100% (7,666), with corresponding performance of 0.276 to 0.717 in F-score. Similar trends can be seen for DDI and CPI also, showing that more data helps improve the performance of supervised CNNs model on all three corpora. But the performance increase slows down significantly once training samples have reached a certain amount (e.g. roughly 40% for both DDI and CPI). We also noticed that CNNs model achieved a higher $F$-score on BioInfer only using 7,666 labeled samples than DDI or ChemProt with ~ 25,000 labeled samples, likely due to the fact that BioInfer is a binary relation corpus, whereas DDI and ChemProt are multi-class relation corpora.

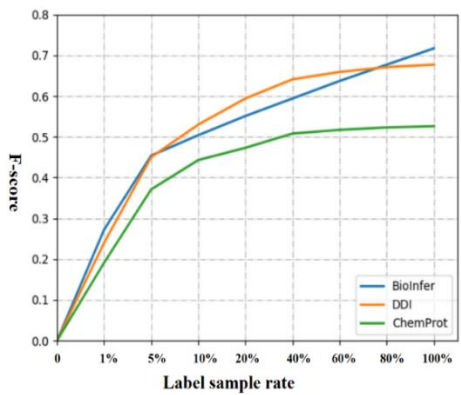

**Fig. 3. The learning curve of CNNs model on three corpora.**

In the biomedical domain, annotating labeled training data is time-consuming and bears high cost. Hence, a critical question is how much labeled data is needed minimally for obtaining competitive performance on biomedical relation extraction. Our results on the three corpora show that the proposed CNNs model with ~2,000 labeled samples can achieve above 70% of the best performance.

Then, we compared our semi-supervised method with the baseline CNN on the same corpora where unlabeled data were directly taken from those un-used instances in the training set (with their labels removed). Table 5 shows that our proposed semi-VAE method can effectively improve the performance of both binary and multi-class relation extractions. For example, on BioInfer, our method improved the $F$-score from 0.483 to 0.544 with 500 labeled data, which outperformed the $F$-score of 0.518 achieved by the CNNs baseline using 1000 l,abeled data. Similarly, on

ChemProt, our method improved the *F*-score from 0.292 to 0.352 using 500 labeled data, which also outperformed the *F*-score of 0.349 achieved by the CNNs baseline using 1,000 labeled data.

**Table 5.** The evaluation results on *F*-score.

| Corpus | Labeled data | CNN | Semi-Supervised VAE |
|---|---|---|---|
| BioInfer | 250 | 0.447 | 0.521 |
|  | 500 | 0.483 | 0.544 |
|  | 1000 | 0.518 | 0.563 |
|  | 2000 | 0.556 | 0.587 |
|  | 4000 | 0.620 | 0.622 |
|  | All (7666) | 0.717 | - |
| DDI | 250 | 0.236 | 0.367 |
|  | 500 | 0.324 | 0.415 |
|  | 1000 | 0.418 | 0.482 |
|  | 2000 | 0.495 | 0.528 |
|  | 4000 | 0.572 | 0.579 |
|  | All (25792) | 0.677 | - |
| ChemProt | 250 | 0.203 | 0.287 |
|  | 500 | 0.292 | 0.352 |
|  | 1000 | 0.349 | 0.381 |
|  | 2000 | 0.418 | 0.443 |
|  | 4000 | 0.505 | 0.509 |
|  | All (25071) | 0.526 | - |

Overall, our semi-supervised method can significantly and consistently improve the performance when the labeled data is 2,000 or less. From Fig. 4, it can be seen also that the performance improvement of our semi-supervised model gradually decreases when more labeled data becomes available. For example, the performance of our method is generally no different to the baseline when 4000 labeled instances were used.

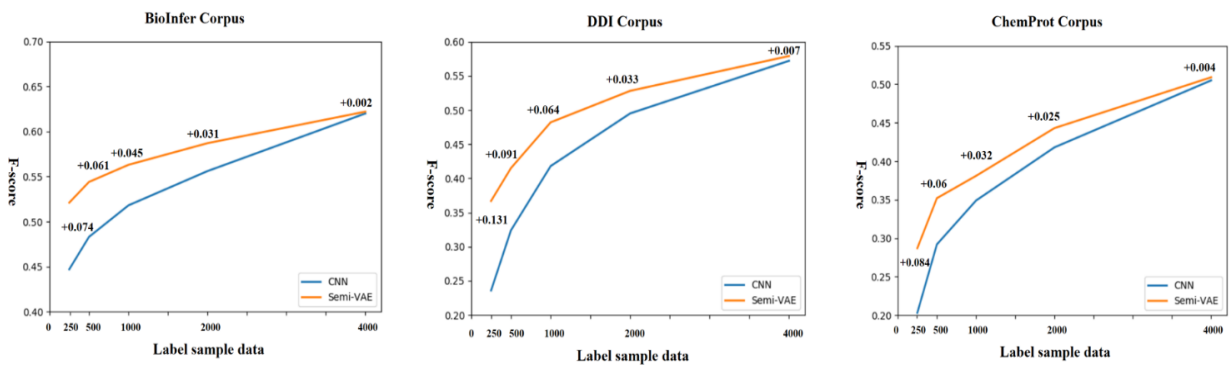

**Fig. 4.** Comparison of baseline CNN vs. Semi-VAE performance on three corpora as the labeled training data is varied from 250 to 4,000.

## 4. Discussion & Conclusions

Automatically extracting relations in biomedical text is a crucial task in biomedical NLP. The supervised-based methods in biomedical relation extraction heavily depend on labeled data, which is expensive and difficult to obtain at a large scale in the biomedical domain. In this work, we propose a semi-supervised VAE-based method to extract biomedical relation while effectively learns useful features from both labeled data and unlabeled data. By exploiting the unlabeled data, our model is able to achieve competitive performance with a small amount of labeled data while taking advantage of freely available unlabeled data in the literature.

In this work, we also experimented and evaluated additional reconstruction strategies, albeit not reported here. For example, we reported above the use of candidate entities surrounding sequences as the encoder input. We also evaluated two other strategies. First, we experimented using the whole sentence sequence as the encoder input, which reconstructs the whole sentence words between the input of the encoder and the output of the decoder. In the other case, we used both the whole sentence sequence and word position information as the encoder input, and we reconstructed not only the words of the whole sentence but also the position of each word. The performance of three reconstruction strategies were found to be similar. However, using candidate entities surrounding sequences was much more efficient than other two strategies because the VAE only needs to reconstruct a small part of the sentence.

Generally speaking, the selection of the labeled data may affect the performance of the semi-supervised method. Thus in our experiments, we randomly selected the labeled data and repeated each training ten times to reduce potential selection bias. Intuitively, the value of each labeled instance may play a different role for the classifier. If we are able to effectively select the best labeled data, it may further benefit our semi-supervised method. Thus, we plan to further investigate how to integrate the active learning strategy into our semi-supervised model in the future.

## Acknowledgments

The authors thank I. Segura-Bedmar, P. Martínez, M. Herrero-Zazo, and T. Declerck for their support with the DDI 2013 corpus.

## Funding

This work was supported by the NIH Intramural Research Program, National Library of Medicine.

## Competing interests

The authors declare no competing interests.